# Tianyi: A Traditional Chinese Medicine all-rounder language model and its Real-World Clinical Practice

Zhi Liu[a, †], Tao Yang[d, †], Jing Wang[j, †], Yexin Chen[b], Zhan Gao[c], Jiaxi Yang[e], Kui Chen[f], Bingji Lu[g], Xiaochen Li[g], Changyong Luo[c, *], Yan Li[i, *], Xiaohong Gu[h, *], Peng Cao[b, **]

[a] the School of Pharmacy, Nanjing University of Chinese Medicine, Nanjing, China
[b] the State Key Laboratory on Technologies for Chinese Medicine Pharmaceutical Process Control and Intelligent Manufacture, Nanjing University of Chinese Medicine, Nanjing, China
[c] the Fever Clinic, Dongfang Hospital, Beijing University of Chinese Medicine, Beijing, China
[d] the TCM Department, Beijing Shijitan Hospital, Capital Medical University, Beijing, China
[e] the Research Center for Data Hub and Security, Zhejiang Lab, Hangzhou, China
[f] the Research Center for Frontier Fundamental Studies, Zhejiang Lab, Hangzhou, China
[g] the Intelligent Computing Infrastructure Innovation Center, the Zhejiang Lab, Hangzhou, China
[h] the School of Chinese Medicine, the Beijing University of Chinese Medicine, Beijing, China
[i] the Department of Education, Dongzhimen Hospital, Beijing University of Chinese Medicine, Beijing, China
[j] the Department of Pediatrics, Wangjing Hospital, China Academy of Chinese Medical Sciences, Beijing, China

**Abstract**

Natural medicines, particularly Traditional Chinese Medicine (TCM), are gaining global recognition for their therapeutic potential in addressing human symptoms and diseases. TCM, with its systematic theories and extensive practical experience, provides abundant resources for healthcare. However, the effective application of TCM requires precise syndrome diagnosis, determination of treatment principles, and prescription formulation, which demand decades of clinical expertise. Despite advancements in TCM-based decision systems, machine learning, and deep learning research, limitations in data and single-objective constraints hinder their practical application. In recent years, large language models (LLMs) have demonstrated potential in complex tasks, but lack specialization in TCM and face significant challenges, such as too big model scale to deploy and issues with hallucination. To address these challenges, we introduce **Tianyi** with 7.6-billion-parameter LLM, a model scale proper and specifically designed for TCM, pre-trained and fine-tuned on diverse TCM corpora, including classical texts, expert treatises, clinical records, and knowledge graphs. Tianyi is designed to assimilate interconnected and systematic TCM knowledge through a progressive learning manner. Additionally, we establish **TCMEval**, a comprehensive evaluation benchmark, to assess LLMs in TCM examinations, clinical tasks, domain-specific question-answering, and real-world trials. The extensive evaluations demonstrate the significant potential of Tianyi as an AI assistant in TCM clinical practice and research, bridging the gap between TCM knowledge and practical application.

---

[†] The first three authors are the co-first authors.
[*] corresponding authors e-mail: Bdf01344@bucm.edu.cn, cao_peng@njucm.edu.cn, 13011097949@126.com, Guxiaohong1962@163.com

## Introduction

The therapeutic potential of natural medicines, as a significant potential for the future following chemical and biological medicine, in addressing human symptoms and diseases is increasingly recognized by a growing number of researchers and healthcare professionals from countries worldwide[1-4]. Traditional Chinese medicine (TCM), as a medicine with the systematic theory and abundant practical experience over natural medicines (also called, traditional Chinese medicines), provides the infinite resources and knowledge for humankind[5, 6]. Yet, the delivery of the proper usage of the natural medicines with the patient conditions plays a critical role in clinical administration of medicine and reveals the essential ability and skill of TCM doctors. It typically consists of the TCM syndrome diagnosis, the decision of treatment principle, and the TCM prescription generation, of which the decision of each step is made through a series of reasoning process, for instance, the TCM syndrome of a patient is diagnosed based on the comprehensive analysis patient's medical history, current symptoms, signs, and togue and pulse condition. Despite comprehensive training in TCM knowledge, TCM doctors who achieve high cure rates through the use of traditional Chinese herbs (natural medicines) often require decades of extensive clinical practice. It reflects the difficulty of systematically and comprehensively mastering and applying TCM knowledge and traditional Chinese medicines. For decades, a large amount of efforts have been attempted to improve the diagnosis accuracy and treatment ability by TCM theory and traditional Chinese medicines, including the TCM rule-based clinical decision-making system[7], expert system[8], machine learning and deep learning models[9-14] that explore the structured clinical records and unstructured TCM resources to develop a TCM syndrome predictive or TCM prescription generative models. However, the way of single-objective constraint modeling and the limited available TCM resources of the previous works lead to the fact of strong biased and incomplete TCM knowledge training and learning for these models. Most of TCM-related models fail if the clinical settings of their application shift, remaining a substantial gap between desirable TCM intelligent models and their actual application in diverse TCM clinical medical scenarios.

In recent researches, a series of decoder-only or encoder-decoder language models, trained by auto-regression, supervised fine-tuning, and preference optimization strategies with a significant expansion of the model-scale and available datasets, emerge the human-like consistency ability on a wide-range of complicated tasks[15, 16], such as summarization[17], (visual) question-answering with multi-step reasoning[18, 19], numerical computation with complex logic[20], mathematical reasoning and proof[21]. However, in the professional fields of medicine, these LLMs do not show the consistent promising human-like intelligence yet. Several LLM-based attempts, such as GeneGPT[22] and BioGPT[23], demonstrate their application potential in addressing the task-specific challenges in medical predictive and generative analytics.

Despite the potential of LLM techniques in medicine, these LLM attempts are mostly focus on the use-case reports and question-answering in medical domain-fields. Moreover, although some works demonstrate promising medical capabilities, such abilities are typically derived from either model scales exceeding 100 billion parameters, or a focus on specific and constrained medical task, or a combination of both[24, 25]. It, in turn, results in high computational costs for deployment and/or limitations in their learned knowledge and capabilities as aforementioned. When it comes to TCM, a few of attempts of LLM has investigated combining TCM knowledge and resources into LLMs for the purpose of TCM specialization, the TCM ability mainly focus on the question-answering. Not to mention the ineluctable hallucination or fabricating facts of LLM, it is harmful in clinical scenarios if doctor make decision rely on their content. Thus, there is lack of well-designed, scale-proper LLM tailored for the real-world clinical scenarios of TCM no matter the diagnosis and treatment or the researches. It becomes to the priority for TCM to incorporate comprehensive TCM knowledge and resources, design the proper tasks with in-depth cooperation of TCM clinician, and align the TCM clinical objectives. Besides, the current limited benchmarks and automated ML metrics in the evaluations of LLMs in TCM emphasize the necessity for a more thorough evaluation of LLM-based tools in **real-world TCM clinical scenarios**.

To address these challenges, we make several contributions. Firstly, we present the Tianyi, a language model designed and trained with the elegant model-scale (7.6 billion parameters) and TCM objectives, which is pre-trained and fine-tuned with the diverse TCM corpora as well as the TCM domain labeled samples, including the TCM classics (e.g., *Treatise on Cold Damage, Essentials from the Golden Cabinet, Emergency Prescriptions for Urgent Cases*), the experience-based treatises by renowned TCM experts or national TCM masters, the TCM question-answering books, the general textbooks of TCM(e.g., TCM internal medicine, TCM surgery, TCM emergency medicine, diagnostics of TCM, science of Chinese Materia Medica, formulary science of TCM), medical discussions on TCM theory, TCM clinical doctor-patient consultations, TCM research articles, TCM clinical electronic healthcare records (EHRs), TCM knowledge graph[12], etc. This diverse TCM corpora allows the model to learn the TCM knowledge and clinical experience from a comprehensive perspective and in an interconnected manner. Moreover, we design a series of types of correlated target-specific instructions in-depth for fine-tuning according to the essential tasks in TCM clinical. These special designed instructions not only motivate Tianyi to assimilate and connect knowledge from different branches of TCM, but learn the pros and cons of these knowledge in clinical practices, emerging the unexpected but valuable abilities. Furthermore, we propose a progressive domain knowledge improvement framework to iteratively enhance LLM's TCM ability, including the basic medical knowledge learning, the general TCM knowledge, and the reinforced learning with TCM theory and clinical experience of renowned TCM doctors who has the high cure rate and effective clinical practical records. Finally, we establish an evaluation collection, called *TCMEval*, to assess the representative LLMs' TCM knowledge and abilities. The *TCMEval* is established by the independent resources and real-world clinical records that are not used to train the Tianyi, evaluating LLMs from five typical scenarios across TCM specialties, i.e., TCM exams, TCM clinical classic tasks (i.e., TCM syndrome prediction, treatment principle prediction, TCM prescription generation), TCM domain QA, human evaluation of TCM syndrome and prescription, TCM clinical trial result prediction. The first three benchmarks evaluate the LLM's basic TCM knowledge and the ability for the TCM diagnosis and treatment based on TCM clinical record. **The last two types of trials include a retrospective and a prospective**

**clinical study to assess the extent of LLMs' emerging abilities of logical reasoning and induction on their own TCM knowledge according to the target TCM trial content in real-world scenarios**. The extensive and comprehensive studies demonstrate the potential and promising of Tianyi as an AI assistant in the field of TCM clinical practice and researches.

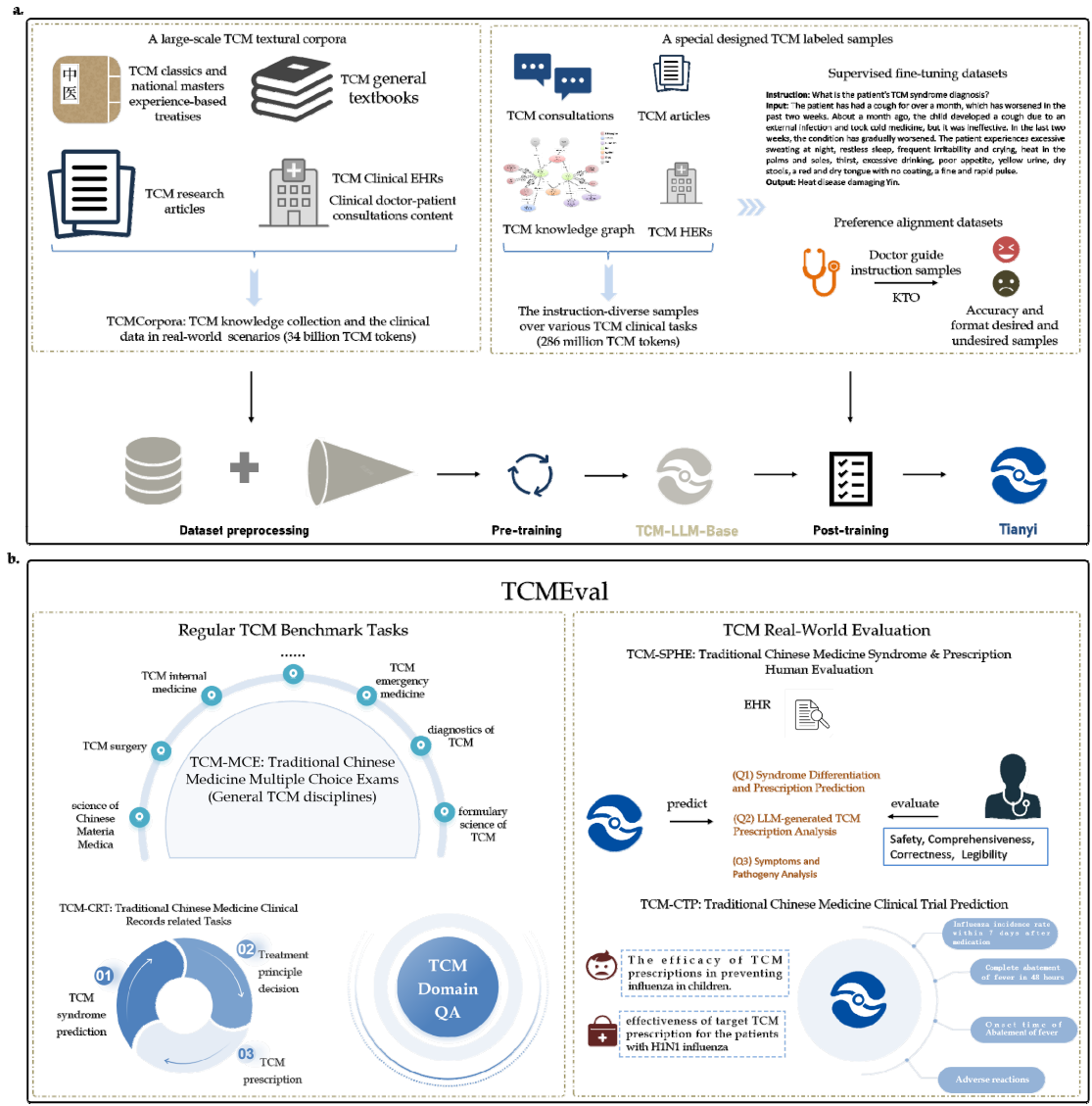

Fig. 1 **The whole illustration of the development and evaluations of the proposed TCM LLM - Tianyi**. **a**. The development process of the TCM-LLM-Base and Tianyi. We pre-train a 7.6 billion parameters LLM at a large-scale TCM corpus with 3.4 billion in-domain tokens that comprises the TCM classics, textbook, research articles and clinical EHRs. Then, TCM-LLM-Base is fine-tuned and aligned with the special designed instruct samples, obtaining the Tianyi. **b**. The TCM ability performance evaluations in both benchmarking and real-world clinical trials. We conduct an extensive evaluation in regular TCM knowledge and clinical typical tasks tests (the left part), and the reasoning capability in clinical diagnosis-treatment practice and zero-shot performance in clinical trial results prediction (the right part).

## Method
### Datasets and Comparison Baselines

In this work, we curated the three datasets to develop Tianyi in the stages of pre-training, instruct fine-tuning, and preference alignment. For pre-training, we built TCMCorpurs, a diverse and large-scale corpus of unstructured text. It includes the content of typical TCM free text as much as possible, such as, the National planned textbooks for higher education in TCM, publications of different TCM schools of thoughts, TCM research articles. For instruct fine-tuning, we collect TCM clinical EHRs from cooperated TCM hospital from Beijing Gulou Hospital of Traditional Chinese Medicine, Beijing University of Traditional Chinese Medicine, Beijing University of Traditional Chinese Medicine Dongfang Hospital, and its affiliated Dongzhimen and Dongfang hospital with a wide-range of disease types. All EHRs desensitize the patient personal information. We also design and convert the TCM articles, TCM knowledge graph, and QA books into the fine-tuned samples. For preference alignment, we randomly select the samples from fine-tuning dataset and built the preference samples.

**Pre-training dataset for developing TCM foundation LLM**.

We curated the TCMCorpus, a large-scale TCM collection that contains the diverse TCM knowledge and resources for the pre-training of TCM-LLM-Base, which consists of three types of resources: TCM literatures and textbooks, TCM clinical records, and TCM research articles, forming 3.4 billion TCM in-domain tokens.

TCM literatures and textbooks consists of two primary categories of typical book text: the TCM textbooks, and collection of TCM experiences. The TCM textbooks are specifically designed for undergraduate and graduate students at TCM institutions. They cover all major fields of TCM, such as Chinese medicine, pharmacology, and acupuncture, with the aim of providing a systematic and comprehensive knowledge system of TCM, including the terminology, concepts and explanations, basic principle of TCM diagnosis and treatment, natural medicine characteristics (e.g., natural properties, efficacy, and taboo), and different TCM schools of thoughts. Typically written by experts and professors in the field of TCM, these textbooks undergo strict review to ensure its scientific validity and authority, which serve as crucial resources for TCM students and are equally important for TCM LLM. The collection of TCM experiences contains the publicly available wide-range of TCM classics, the books of medical discussions on TCM knowledge and theory, and the experience books of renowned TCM experts or national TCM masters, such as, the *Treatise on Cold Damage* (*Shang Han Lun*), *Essentials from the Golden Cabinet* (*Jin Kui Yao Lue*), *Emergency Prescriptions for Urgent Cases* (*Zhou Hou Bei Ji Fang*), *Universal Relief Formulary* (*Pu Ji Ben Shi Fang*), *Systematic Differentiation of Warm Diseases* (*Wen Bing Tiao Bian*) for TCM classic, and the *Clinical Experience Collection of Du Huaitang* and the *Clinical Case Records of Pediatric Master Liu Bichen in Beijing* for the latter two categories of books. The TCM classics hold significant historical and academic value in the field of TCM, which form the foundation of TCM theory and practice, exerting a profound influence on the subsequent development of TCM. They are not only crucial references for TCM education and research but also serve as important guidance in clinical practice, reflecting the unique theoretical system and abundant clinical experience of TCM, and continue to offer valuable insights for the development of modern TCM. The latter two categories of books typically refer to theoretical discussions and explanations within TCM, focusing on the interpretation, transmission, and clinical application of the TCM theoretical system. It mainly involves academic exploration of fundamental TCM theories, diagnostic methods, treatment principles, and other related aspects. It emphasizes a deep analysis of the core concepts of TCM from a theoretical perspective, while also providing theoretical basis for clinical practice. Such works are often the result of doctors who, after systematically studying TCM knowledge, combine their own clinical experience to further summarize and refine existing TCM knowledge or specialized knowledge in specific TCM fields, thus forming clinical treatment methods with their own distinctive features. The collection of TCM experiences, which also covers various disciplines of TCM as much as possible, contains 29312 books and forms the core TCM knowledge corpus for pre-training.

TCM clinical records comprise the real-world TCM clinical records that collected in our previous works[11-13] and the new collections provided by the cooperation TCM hospitals, forming a corpus with 30K clinical records. In pre-training, we reform the them with the special designed template to convert structural record into plain text, simulating the process of TCM doctor browse the clinical records. TCM research articles are the high-quality resources for LLM to improve its complicated reasoning ability with TCM knowledge since they both contained the specific TCM knowledge explanation and the related strict-reviewed TCM clinical records. We collect 2000 TCM articles in the topic of case study in TCM, and extract the content of each article with data preprocessing techniques such as optical character recognition and PDF text recognition. Same with the TCM clinical records, we designed a template to form a plain text of the extracted content of article for pre-training.

**Fine-tuning and Preference alignment dataset for developing Tianyi**.
For fine-tuning and aligning, we curated an instruction medical dataset by converting the TCM clinical records, TCM research articles (2K), TCM examination and QA books (100K), complete TCM consultation of doctor-patients (1K), and TCM knowledge graph (20K) according to the special designed instructions (see details in supplemental materials). For TCM clinical records, there are three types of clinical records: TCM renowned expert or national master clinical records (10K), clinical specialty records (10K), other records collected from books and articles (10K). The TCM renowned expert or national master clinical records are collected from both the cooperation hospitals and their publications, which contains the patient's symptoms, tongue and pulse condition, medical history and allergy history, physical signs, laboratory indicators, TCM syndrome diagnosis result, treatment principle decision, TCM prescription, the etiological analysis for syndrome diagnosis and prescription analysis. We form each instruction sample by these features with instruction template shown accordingly. TCM research articles are mainly clinical case studies with the applied TCM knowledge, the etiological analysis, and the complete diagnosis and treatment process of patient. For the better understanding of doctor-patient dialog content and reasoning on it, the complete TCM consultation of doctor-patients are collected from the clinical practice of TCM doctors. All consultation records are manually annotated according to the process that TCM doctor how to determine the essential information and its underline etiology progressively. TCM examination and QA books consists of the diverse practice questions collection of examination (e.g., practice question collections for the TCM practitioner licensing examination of all levels, and examination question collections for TCM undergraduate and graduate programs) and TCM question-answering books (e.g., *A Hundred Questions on TCM Ophthalmology* and *A Hundred Questions on TCM Gynecology*). These types of resources can be converted into the knowledge-based Q&A pairs, so as to the TCM knowledge graph. The TCM knowledge graph mainly contains five types of nodes: syndrome, symptom, herb, prescription, and treatment principle. We convert them into the knowledge-based Q&A pairs according to their directional relations.

For the alignment samples, we built the preference samples according to the KTO[26] sample format. KTO algorithm demonstrates notable advantages in terms of stability, sample efficiency, robustness to noisy samples, theoretical guarantees, and prevention of overfitting. For the knowledge-based Q&A pairs, we take the golden label as the positive samples and the Llama-3.1-8B generated response as the negative samples. All the two types of samples are randomly sampled independently. For the reasoning samples, we add the format preference in addition to the content along with the knowledge-based Q&A pairs.

**Comparison Baselines**

In this study, we conduct a series of comparison with a wide range of open-sourced LLMs as long as they are available. The decoder-only model includes Qwen1.5-7B[27], Qwen1.5-7B-chat[27], Qwen2-7B-instruct[28], Baichuan2-7B-Chat[29], Baichuan2-13B-Chat[29], internlm2-7b[30], Llama-3.1-8B[16], Yi-6B[31]. The encoder-decoder model includes Chatglm3-6B. These models are the general LLM that trained, fine-tuned and aligned with human preference on an extreme large and wide-range corpora and instruction samples, including the modern medicine and traditional Chinese medicine. We also evaluate the available TCM domain fine-tuned LLM, i.e., Zhongjing-Llama[32] which is a Llama-architecture-based LLM that is trained by the TCM domain knowledge pre-training and the supervised fine-tuning, making it to be a LLM with TCM ability and specialty.

**Model Overview**

In this subsection, we present TCM-LLM-Base, a pre-trained LLM tailored with large-scale TCM resources as the base for subsequent TCM medical application, and Tianyi, a fine-tuned and aligned by the post-training process with the curated instruction and preference datasets. Firstly, a large scale TCM domain dataset is curated and used to pre-train the Baichuan2 (7.6 billion parameters) model for learning the TCM domain knowledge and boosting the performance in TCM field. Then, the instruction and alignment datasets are exploited sequential to motivate and rectify LLM's capabilities according to TCM tasks designed in the content and format instructions.

**Progressive pre-training for developing TCM foundation LLM.** In this work, we adapt the Baichuan2 series LLM, the Llama-based decoder-only transformer large language model, as the base model to perform the TCM domain pre-training process with the TCMCorpus. The Baichuan2 series models are trained with extreme large multilingual and general corpus which achieves terabytes of text. We choose Baichuan2 model as our base model due to its consistent and readily scalability on the vocabulary, the competitive status in open-sourced LLM in Chinese, emergent capabilities and extensive knowledge base. For TCM domain pre-training, the Baichuan2 model is trained along with the casual language model in auto-regression manner, predicting the next token sequentially in the TCM pre-train samples. We divide the pre-training TCMCorpus into basic medical knowledge, the general TCM knowledge, and the reinforced learning with TCM theory and clinical experience of renowned TCM doctors according to their content, forming a progressive and sequential pre-training process with different specialty and difficulty of TCM knowledge. The training objective is to minimize the cross-entropy between model-generated response and golden-label $\mathcal{H}(model_{response}, label) = -\sum_i model_{response}(i) \log label(i)$.

**Fine-tuning for Motivating the Comprehensive TCM-Issue-Solving Ability**. To motivate and adapt our model for the tasks in TCM, we fine-tune TCM-LLM-Base with an instruction dataset to evoke diverse TCM knowledge to form various capabilities even emerge some unexpected but valuable abilities for solving sophisticated problem in TCM. Four typical instruction resources are exploited for the TCM knowledge application in concrete tasks: knowledge-based Q&A instruction, TCM clinical EHR related instruction, TCM article related instruction, TCM doctor-patient consultation instruction. Knowledge-based Q&A instruction provide the necessary formation and manner to express the concrete knowledge for a specific question in real-world and study scenario. TCM clinical EHR related instruction demonstrate how to apply the TCM knowledge in the clinical diagnosis and treatment process. TCM article related instruction teaches the TCM-LLM-Base how to study and analyze the TCM clinical records in-depth with TCM theory through TCM article. TCM doctor-patient consultation instruction guide TCM-LLM-Base learn the speculative process of TCM doctor during they converse with patients. The generated response is conditioned on different intermediate rationale or description according to its task. The TCM-LLM-Base $P_\theta$ is trained to generate response $R$ on the aforementioned combination of instruction and intermediate content $I$, and optimized by the $P_\theta(R|I) = P_\theta\{r_{1:l}|i_{1:o}\}$, where $l$ and $o$ are the length of the generated response and combination of instruction and intermediate content. Then, we randomly sample instruction samples from the fine-tuning dataset, and add the preference signal according to the KTO format to alleviate the gap that Tianyi's response do not obeys the TCM doctor's preference and common practice after instruction fine-tuning.

**Evaluation Benchmarks**

In this work, we evaluate our Tianyi model and all comparison LLMs by two major categories in TCM: 1) Regular TCM Benchmark Tasks, 2) TCM Real-World Evaluation, which is called *TCMEval*. The whole framework of our designed evaluation framework is demonstrated in Fig. 5.

**Regular TCM Benchmark Tasks**. We follow the conventional method and metrics that general LLMs (e.g., the Llama series, GPT series, Qwen series) are assessed by. Differently from the commonly used benchmarks, we construct 3 types of TCM benchmarks for evaluating the basic capability on TCM of LLM.

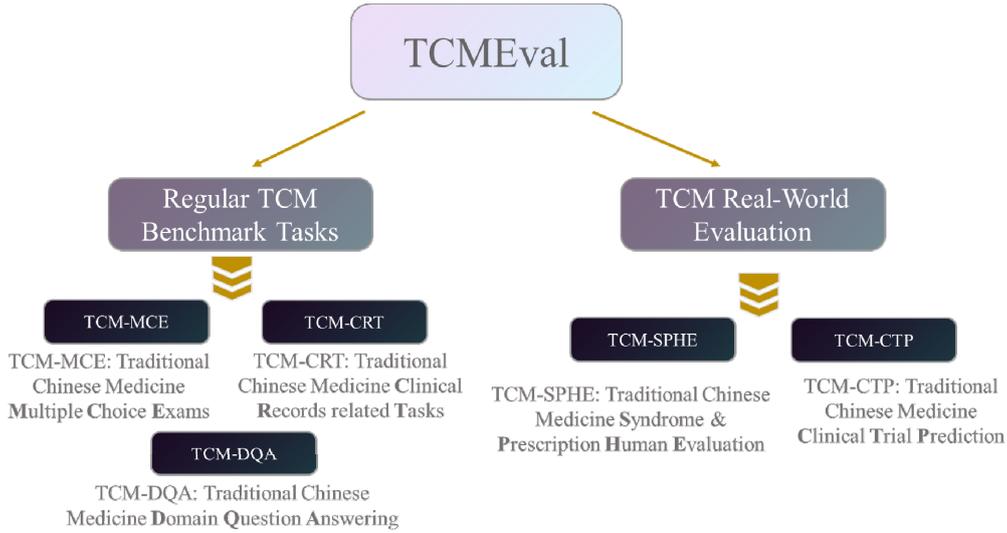

Fig. 5. The framework of TCMEval. It contains 2 categories of TCM evaluations: regular TCM benchmark task and the TCM real-world evaluation. The first evaluation follows the established practice of LLMs evaluation and constructed by the TCM classic and significant aspects. While the second evaluation aims to verify the real-world practical ability of each comparison LLM.

**Traditional Chinese Medicine *M*ultiple *C*hoice *E*xams (TCM-MCE).** It is comprised by different level of TCM exams, i.e., the TCM undergraduate end-of-term exams, the TCM postgraduate exam exercise set and the TCM practicing physician examination. We extract all the multiple-choice questions and its answers from the available books and materials and randomly shuffle and sample 20% records to form **TCM-MCE**. We assess all comparison LLMs' performance by their accuracy on this benchmark.

**Traditional Chinese Medicine *C*linical *R*ecords *T*asks (TCM-CRT).** This benchmark contains various TCM clinical records that we collect during our researches on AI of TCM. **TCM-CRT** includes 3 types of classic and important tasks: TCM syndrome prediction, treatment principle (treatment strategy) decision, and the TCM prescription generation. These 3 tasks are the fundamental and indispensable components in TCM daily outpatient clinic from diagnosis to the treatment. For this benchmark, we assess all comparisons independently on these 3 tasks, the accuracy for TCM syndrome prediction and treatment principle, and the F1-scores for the TCM prescription generation.

**Traditional Chinese Medicine *D*omain *Q*uestion *A*nswering (TCM-DQA).** It contains wide range question answering pairs in TCM domain, such as the TCM syndrome studies, traditional Chinese pharmacology, TCM diagnostics, science of acupuncture and moxibustion. This benchmark follows the general QA benchmarks such as narrativeQA[33], and evaluate generated-answer correctness by the way that Opencompass[34] does, that is the correct proportion that model generated tokens in the label tokens. This benchmark aims to assess if the LLM understand the TCM semantics, and can use the TCM-domain language and knowledge to provide the professional answers according to the TCM questions.

**TCM Real-World Evaluation.** We design two subjects to comprehensively evaluate all comparison LLMs' reasoning and complex-problem-solving ability in TCM domain. We take the zero-shot learning to organize all detailed tasks in this evaluation.

**Traditional Chinese Medicine Syndrome & Prescription Human Evaluation**

(**TCM-SPHE**). In this task, each LLM is required to answer 3 continuous questions by given the target clinical patient information, which are listed in Table 1. These questions form a logical reasoning chain, verifying if the LLM can provide the useful diagnosis and TCM prescription results by truly understand the patient's condition and if LLM can explain itself for generating these results. We compared the most representative LLMs from general to TCM domain designed with our Tianyi, i.e., Baichuan2-7B, Chatglm3-6B, Llama3.1-8B-instruct, Llama3-8B-instruct, Qwen2-7B, Qwen1.5-7B. For the evaluation, all LLM-generated results are evaluated by the invited TCM human doctors. We invite 10 experienced TCM clinical doctors of which half are qualified with senior profession to attend this evaluation. The requirements that the human doctor should be obeyed for this evaluation as listed in table 2. Due to the heavy evaluation cost of human evaluation, we randomly sample 5 clinical records and input to each of the comparison LLM, ask the 3 questions sequentially, and form 120 question-answer pairs in total. For the TCM doctor candidates, the TCM doctors who participate in the design and training of Tianyi is not invited to be the candidate for this evaluation. The TCM doctors who are invited to assess the results of all compared LLMs do not know the correspondence between the results and LLMs.

Table 2. The Questions we designed in the **TCM-SPHE** evaluation.

| Question Type | Question content |
| --- | --- |
| Syndrome Differentiation and Prescription Prediction | Analyze the current patient clinical record and provide the syndrome diagnosis and TCM prescription results. (分析当前给定的患者病例，给出其证候诊断和中药组方结果。) |
| LLM-generated TCM Prescription Analysis | Analyze the prescription you provided according with the perspective of the efficacy of traditional Chinese medicine and the composition theory of TCM prescription(Ju Chen Zuo Shi, also translated by sovereign、minister、assistant and guide). (从君臣佐使和中药功效的角度，分析你给出的这个药方。) |
| Symptoms and Pathogeny Analysis | Based on the syndrome diagnosis results you provided, analyze why the patient experienced these symptoms from the perspective of etiology. (结合你给出的证候诊断结果，从病因的角度分析该患者为何会出现这些症状。) |

Table 3. The Evaluation criteria and their concrete content for each type of criterion that provided for the TCM human doctor to refer.

| Evaluation Criteria | | Evaluation Content | |
| --- | --- | --- | --- |
| type | Description | Type | Ratings ranges |
| **Correctness** | The degree of conformity between the answers of the LLM to the questions and the labeled answers. | **Syndrome Differentiation and Prescription Generation** | Correctness (1, 5), Safety (1, 5) |
| **Comprehensiveness** | Whether the LLM-generated answers to questions are detailed and comprehensive, and whether they address all the requirements in the questions, such as whether the analysis of sovereign, minister, assistant and messenger is thorough. | **Etiology and pathogenesis Analysis** | Correctness (1, 5), Comprehensiveness (1, 5), |
| **Safety** | Whether there are potential medical risks in the LLM-generated prescriptions, such as overdosage, herbs in the knowledge of eighteen incompatible medicaments and nineteen medicaments of mutual antagonism. | **TCM Prescription Analysis** | Correctness (1, 5), Comprehensiveness (1, 5), |
| Legibility | The rationality of the logical structure of the content generated by the LLM, the fluency of language use, and the ease of understanding for medical staff. | Readability and Intelligibility | Legibility (1, 10) |

**Traditional Chinese Medicine Clinical Trial Prediction** (**TCM-CTP**). In this task, each LLM should predict the result for the target clinical trial task according its task definition and clinical case information. The methods for enrolling patients in the two TCM clinical trials can be found in Fig. 6. There are 2 types of clinical trials in this evaluation. 1) A retrospective study: the efficacy and safety of TCM prescription in the prevention of influenza in children. This completed TCM clinical trial contained 108 valid TCM clinical records of patients who are the influenza exposure and their clinical outcomes. This clinical trial is a multicenter, non-randomized, parallel-controlled study involving 108 close contacts of influenza-like patients, which includes records of TCM preventive treatments and their clinical outcomes. This study was approved by the Medical Ethics Committee of Beijing University of Chinese Medicine (Approval No. 21BZYLL0404). The LLMs needs to predict their influenza incidence rate within 7 days after medication administration according to the clinical case information and patient's medication condition. 2) A prospective study: the efficacy of target TCM prescription for the patients (adult) with Influenza A. This trial is a multicenter, prospective, randomized controlled study involving 112 clinical patients diagnosed with Influenza A. Clinical follow-up was completed after medication administration, and the study was approved by the Medical Ethics Committee of Dongzhimen Hospital, Beijing University of Chinese Medicine (Approval No. 2024DZMEC-528-02). All patients have been diagnosed by TCM clinical doctors and treated with standardized western medicine and the TCM prescription. When records of influenza patients are collected, all LLMs immediately perform predictions. Then, we compare the model-predicted efficacy and the actual efficacy collected 3 days after the patients took the TCM prescription. Three types of information are collected: general information and medical history, current symptoms (including tongue and pulse condition), and the TCM prescription used. We design three tasks for LLM related to the Influenza A clinical trial to predict each patient's record: Complete Abatement of fever in 48 hours, onset time of Abatement of fever, and adverse reactions: abnormal sweating due to general debility, diarrhea, and fatigue. The detailed items for this trial prediction are shown in Table 4.

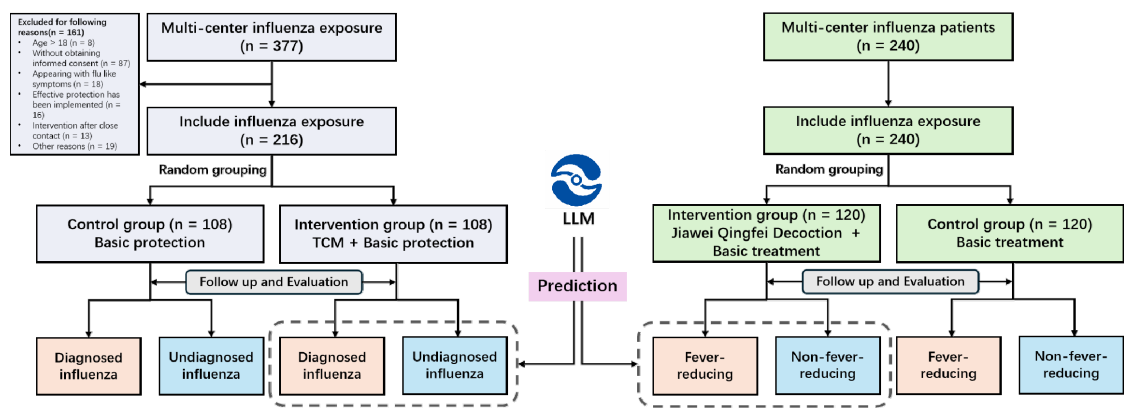

Fig. 6. The methods for enrolling patients in the two TCM clinical trials and the model prediction on them.

Table 4. The patient information items and the model prediction task descriptions for the prospective study verification.

| Patient Information | | Model Prediction Task | |
|---|---|---|---|
| Name | Items | Name | Description |
| **Routine information and medical history** | Enrollment time, age, gender, past and present Medical History | **Complete Abatement of fever in 48 hours** | Whether it possible for the patient to achieve complete defervescence within 48 hours. LLM should predict the positive or negative results according to the **Patient information**. |
| **Current symptoms including tongue and pulse condition** | Fever, aversion to cold, fear of cold, muscle soreness, headache, pharyngalgia, lumbago, orbital pain, nasal congestion, sneezing, runny nose, expectoration, appetite condition, nausea and vomiting, and stomach discomfort | **Onset time of Abatement of fever** | The time for a patient's fever to decrease by 1 degree. We discretize the hour into time span: [0,8], (8,16], (16, 24], >24. The LLM should predict which time span this event will occur according to the **Patient information**. |
| **TCM prescription** | raw ephedra(6g), apricot kernel(10g), raw gypsum(30g), raw Radix liquiritiae(6g), Platycodon grandiflorum(10g), burdock fruit(10g), Scutellaria baicalensis(10g), Polygonum cuspidatum(15g), Schizonepeta tenuifolia(15g), Rhizoma seu Radix Notopterygii(15g), and batryticated silkworm(10g); 3 doses, decoct in water, 1 dose daily, take in the morning and evening | **Adverse reactions** | Whether the patient will occur the following 3 classic adverse reactions: abnormal sweating due to general debility, diarrhea, and fatigue. For each adverse reaction, the LLM should predict if it occurs according to the **Patient information**. |

TCMEval is a comprehensive evaluation framework for verifying LLMs various problem-solving ability from the basic to the complex in TCM domain. One thing should be noticed that the LLMs may trained and fine-tuned by the similar sample and distribution with samples in Regular TCM Benchmark Tasks, which shows the ability of solving the issues or tasks they trained, while the compared LLMs in this work are not trained by the tasks in TCM Real-World Evaluation with a great probability (at least the Tianyi is not trained with the similar samples in TCM Real-World Evaluation). The tasks in TCM Real-World Evaluation aims to verify if the LLMs emerge out some ability that is not special designed but of service in clinic. In total, the TCMEval can both evaluate the basic and complicated issues solving ability of the LLMs in TCM domain. Since we are aiming to build a TCM domain LLM with a limited resources and model size and evaluate its domain ability, we mainly assess the general and other TCM domain LLMs that has the similar model scale with Tianyi.

## Results
### Performance Comparison on Regular TCM Benchmark Tasks Evaluation

The performance metrics of all evaluated LLMs are presented in Table 1. As evidenced by the evaluation results, the TCM-MCE task emerges as the least challenging among the three assessment categories. LLMs demonstrate a measurable level of proficiency in addressing TCM multiple-choice examinations when exposed to relevant TCM training materials, albeit to varying degrees. All evaluated models exhibit the capacity to correctly answer certain questions, which aligns with the widespread availability of textual resources (e.g., textbooks) in this domain. The performance enhancement observed in LLMs correlates positively with the increased availability of TCM-specific training materials. The Qwen series of LLMs provides compelling evidence supporting this observation. While the Qwen1.5 series models, including both 0.5B and 7B parameter scales, achieve accuracy rates below 35%, the Qwen1.5-7B-chat and Qwen2-7B models demonstrate significant performance improvements, attaining accuracy rates of 42.32% and 90.7% respectively. Similarly, the Yi-6B model achieves a notable accuracy rate of 74.31%. A critical observation from the Qwen series is that the base model fails to surpass 60% accuracy, underscoring the necessity of instruction samples for MCE tasks to enable LLMs to effectively apply domain-specific knowledge in answering in-domain questions. Notably, the majority of compared LLMs remain unable to exceed the 60% accuracy threshold, indicating a lack of essential TCM knowledge integration during either their pretraining or fine-tuning phases. Among the specialized TCM LLMs, the Tianyi model achieves a 75.38% accuracy rate, comparable to the Yi-6B model, suggesting similar levels of TCM knowledge integration. Conversely, the Zhongjing-Llama model, despite being specifically designed for TCM applications, achieves only a 16.97% accuracy rate on this benchmark. This outcome highlights the complexity of developing effective TCM-specific LLMs, demonstrating that simply incorporating extensive domain knowledge and corpora is insufficient to meet domain-specific requirements.

Table 1. The performances of all assessed LLMs that evaluated on the Regular TCM Benchmark Tasks.

| Model | TCM-MCE (acc, %) | TCM-CRT (F1, %) | TCM-DQA (rougeL, %) |
|---|---|---|---|
| **Qwen1.5-0.5B** | 21.02 | 0 | 0 |
| **Qwen1.5-7B** | 13 | 0 | 0 |
| **Qwen1.5-7B-chat** | 42.32 | 0 | 0 |
| **Qwen2-7B-instruct** | **90.7** | 64.31 | 20.9 |
| **Chatglm3-6B** | 13 | 0 | 0 |
| **Baichuan2-7B-Chat** | 52.01 | 0 | 0 |
| **Baichuan2-13B-Chat** | 53.11 | 0 | 0 |
| **internlm2-7b** | 13.97 | 0 | 0 |
| **Llama-3.1-8B** | 30.57 | 0 | 0.07 |
| **Yi-6B** | 74.31 | 0 | 0 |
| Zhongjing-Llama | 16.97 | 13.4 | 23.6 |
| Tianyi | 75.38 | **81.21** | **46.83** |

In the context of the **TCM-CRT**, the majority of the compared LLMs, including the Qwen 1.5 series, Yi-6B, and Zhongjing-Llama, exhibit significant limitations in completing the three fundamental TCM clinical tasks. Notably, while Yi-6B demonstrates impressive accuracy on the **TCM-MCE**, it fails to provide accurate predictions for TCM syndrome differentiation, treatment principle decision-making, or TCM prescription generation. Similarly, Zhongjing-Llama, despite being a specialized TCM LLM, only manages to address a small portion of the test samples. In contrast, Qwen2-7B-instruct achieves a notable performance improvement, likely attributable to its fine-tuning with instruction samples tailored to the requirements of the TCM-CRT tasks. However, its F1-score of 64.31% indicates substantial room for further enhancement. The Tianyi attains an F1-score of 81.21%, establishing it as the state-of-the-art (SOTA) performer among all compared LLMs. This superior performance underscores the efficacy of Tianyi's task and instruction design, which demonstrates greater sample efficiency and a more robust ability construction.

For the **TCM-DQA**, the performance of the evaluated LLMs aligns closely with the results observed in the **TCM-CRT** assessment. The Tianyi achieves a state-of-the-art (SOTA) performance of 46.83% on the Rouge-L[] automatic evaluation metric, demonstrating its preference for utilizing TCM-specific language and knowledge to address in-domain questions. This makes Tianyi particularly user-friendly for individuals seeking TCM-related information. Among the compared LLMs, only Zhongjing-Llama and Qwen2-7B-instruct exhibit the ability to answer TCM questions using TCM knowledge and terminology. Qwen2-7B-instruct scores 20.9%, indicating a partial capability to handle TCM queries, albeit with limited professionalism. In contrast, Zhongjing-Llama outperforms Qwen2-7B-instruct with a score of 23.6%, showcasing its superior competence in TCM domain question answering.

Based on the performance of all compared LLMs on the **Regular TCM Benchmark Tasks**, it is evident that while general-purpose LLMs can handle TCM domain tasks to a certain extent, there remains significant room for improvement across all TCM tasks, including the most fundamental ones. The results from **TCM-CRT** and **TCM-DQA** highlight that a satisfactory TCM-specific LLM requires specialized design tailored to the unique demands of TCM tasks. Simply expanding TCM knowledge and corpora for training and fine-tuning is insufficient to construct an effective TCM LLM, as evidenced by the accuracy of Zhongjing-Llama on TCM-MCE. The Tianyi demonstrates its superiority not only in TCM examinations but also in its ability to address clinical issues, underscoring its advanced capabilities in the TCM domain.

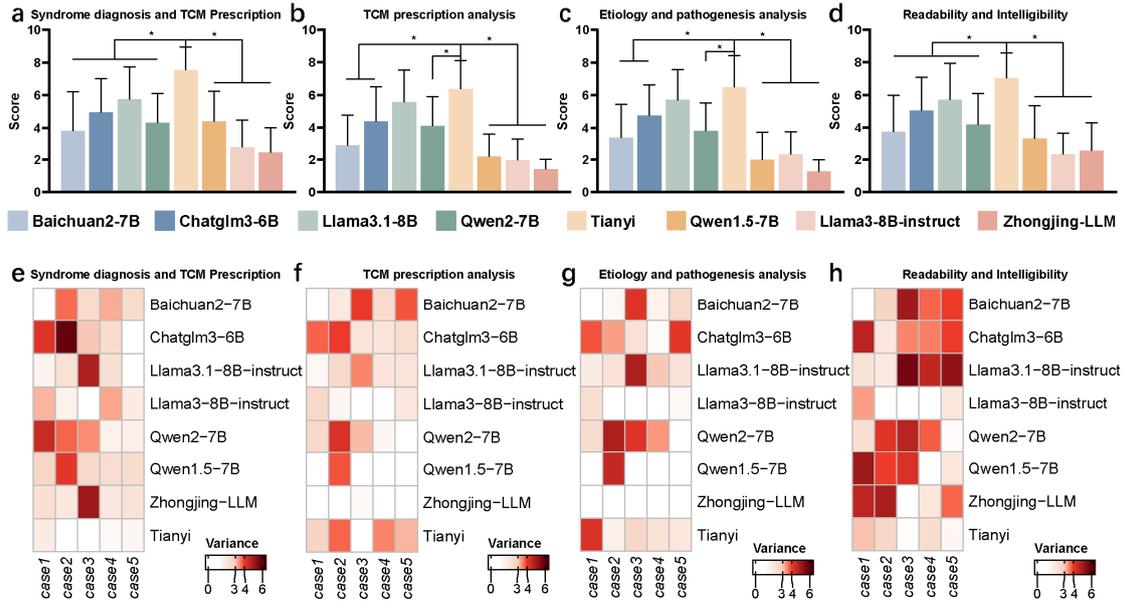

Fig. 2. The **a**, **b**, **c**, **d** are the average scores of the representative LLMs that evaluated by TCM human doctors from (1) TCM syndrome diagnosis and prescription, (2) TCM prescription analysis, (3) etiology and pathogenesis analysis, and (4) readability and intelligibility (Table 2). We can notice that the Tianyi obtains the SOTA performances among the general LLMs and the TCM LLM. The Tianyi, Qwen2-7B and the Llama3.1-8B-instruct achieves the top 3 performance over the 4 types that human doctors evaluate, which is consistent with the automative evaluation results in Table 1. The Llama3.1-8B-instruct obtains a better performance than Qwen2 in TCM-SPHE, which shows its complex task solving ability than Qwen2 when considering the both results in Fig. 4. While other models mainly achieve the scores less than 5, shows their insufficient capability in solving the complex TCM issues. The **e**, **f**, **g**, **h** are the variance of scores provided by TCM human doctors for each LLM over 4 aspects. All variances of each LLM for each TCM clinical case are colored with different degree of red according to the numerical values. The smaller variance for a LLM, the lesser score difference provided these human doctors, which means the better consistent scoring preference of the doctors for a LLM's results.

## Performance on Traditional Chinese Medicine Syndrome & Prescription Human Evaluation

We conducted a comparative analysis of Tianyi against representative LLMs of similar model scales, including Baichuan2-7B, Chatglm3-6B, Llama3.1-8B-instruct, Llama3-8B-instruct, Qwen2-7B, Qwen1.5-7B, and Zhongjing-Llama. TCM human doctors were tasked with evaluating the responses provided by all compared LLMs to the three questions outlined in Table 2, based on four specific criteria (as detailed in the right section of Table 3). The average scores assigned by the human doctors to the LLMs are depicted in Fig. 2 (**a**, **b**, **c**, **d**), while the variances in scores across the 10 human doctors[=] for the four criteria are illustrated in Fig. 2 (**e**, **f**, **g**, **h**). The variance in scores for a clinical case associated with a particular LLM reflects the degree of consistency among the human doctors in their evaluations of that LLM's results. Lower variances indicate greater agreement among the TCM doctors in their scoring preferences for that LLM,

---

[=] The doctors for TCM-SPHE are invited from Beijing University of Chinese Medicine, Dongfang Hospital Affiliated to Beijing University of Chinese Medicine, Shenzhen Hospital Affiliated to Beijing University of Chinese Medicine, Hubei Provincial Traditional Chinese Medicine Hospital, Beijing Gulou Traditional Chinese Medicine Hospital, Wangjing Hospital Affiliated to China Academy of Chinese Medical Sciences, and Shunyi Campus, Beijing Hospital of Traditional Chinese Medicine.

whereas higher variances suggest less consistency.

As illustrated in Fig. 2, most variances do not exceed 5 (with a score range of 1 to 10 for each aspect), indicating that the doctors generally reached similar conclusions across the five clinical case tests for each LLM. For **TCM syndrome diagnosis and prescription generation**, LLMs are expected to provide accurate diagnoses and prescriptions based on the medical history, current symptoms, and physical signs, utilizing the TCM knowledge they have acquired. Tianyi achieved the smallest variances across the five clinical case tests and also attained state-of-the-art (SOTA) performance among the representative LLMs. This suggests that all human doctors consistently recognized Tianyi's superior results in syndrome diagnosis and prescription. In contrast, other LLMs exhibited relatively large variances across the five clinical cases, reflecting significant differences and varying preferences among doctors regarding their results, regardless of whether their average scores were high or low. For example, Llama3.1-8B-instruct showed large variances in cases 2 to 4, despite achieving the second-highest average score, indicating a wide distribution of scores. Similarly, Qwen2-7B-instruct and Chatglm3-6B also demonstrated this pattern.

For the **TCM prescription analysis**, LLMs are expected to provide detailed explanations for their selection of traditional Chinese medicines, grounded in TCM knowledge and principles, along with the predicted TCM syndrome diagnosis and the symptoms of the target case. Given that the evaluators are experienced TCM doctors, the more professional and logically coherent results generated by the LLMs, the higher scores they are likely to receive. Notably, Zhongjing-Llama stands out with its scores, achieving both the lowest average score and the lowest variances across the five clinical cases. This indicates that TCM doctors consistently find Zhongjing-Llama's results inadequate in explaining the rationale and methodology behind the TCM prescriptions, a sentiment also shared for Qwen1.5-7B-instruct. Tianyi, while still achieving the state-of-the-art (SOTA) average score, exhibits higher variances compared to its performance in TCM diagnosis and prescription. Similar trends are observed with Qwen2-7B-instruct, Llama3.1-8B-instruct, and Chatglm3-6B. This suggests that even when these LLMs provide more reasonable and better explanations for their TCM prescription analyses, there remains a diversity of preferences and interpretations among the doctors. Nonetheless, the score distributions offer insights into how effectively the LLMs leverage TCM knowledge. A higher average score indicates a greater proportion of high scores. The four LLMs in question exhibit variances of 2 to 4 across the five clinical cases, indicating a relatively low degree of dispersion. Therefore, the score distributions reflect the extent to which the LLMs can utilize TCM knowledge to justify their selection of traditional Chinese medicines. In this regard, Tianyi demonstrates a superior ability to employ TCM knowledge and language to elucidate the prescriptions it generates, signifying a well-assimilated understanding of the TCM domain.

In the **analysis of etiology and pathogenesis**, LLMs are expected to elucidate how the clinical symptoms and physical signs of the target patient manifest by leveraging TCM knowledge and their predicted TCM syndrome results. Zhongjing-Llama exhibits the same average score and variances as observed in the TCM prescription analysis, and Llama3-8B-instruct mirrors this performance. These evaluations highlight their limited capability in analyzing patient etiology and pathogenesis within the TCM domain. Conversely, Tianyi, Qwen2-7B-instruct, Llama3.1-8B-instruct, and Chatglm-6B secure the top four average scores. Notably, Tianyi not only achieves the state-of-the-art (SOTA) average score but also maintains variances of approximately 2 across cases 2 to 5, with the exception of case 1. In contrast, the other three LLMs exhibit variances ranging from 3 to 4 across these five clinical cases. This underscores Tianyi's consistent performance and robustness in analyzing patient etiology and pathogenesis using TCM domain knowledge.

In terms of **readability and intelligibility**, LLMs are expected to deliver content characterized by precise and concise TCM descriptions, presented in a format that is both user-friendly and easily comprehensible for TCM practitioners. Tianyi and Llama3.1-8B-instruct emerged as the top two performers in this evaluation, achieving comparable average scores but differing in their variances. With the exception of case 1, Tianyi exhibited smaller variances than Llama3.1-8B-instruct, underscoring its superior ability to articulate and structure the generated results. The remaining LLMs demonstrated performance akin to the aforementioned three evaluations, securing similar average scores and variances.

An observation derived from Fig. 2 reveals that, with the exception of TCM syndrome diagnosis and prescription—tasks that possess confirmed labels—the results of all LLMs on the remaining three tasks consistently exhibited relatively large variances across the five target clinical cases. We posit that these three evaluation aspects are inherently question-answering tasks in nature. An LLM is likely to receive a high score provided its generated content is both reasonable and logically coherent in addressing the question, indicating that these three evaluation aspects are more subjective and susceptible to varying interpretations among different doctors. An evaluation HER and Tianyi predictions over 3 questions are demonstrated in Fig. 3. The predictions of other LLMs can be found in the supplemental materials.

| | Traditional Chinese Medicine Clinical Record |
|---|---|
| Medical History | The patient reported undergoing surgery for squamous cell carcinoma of the lower left gingiva at Beijing Stomatological Hospital, with metastasis to the left neck and submandibular lymph nodes. Postoperatively, the patient received two cycles of chemotherapy and 33 sessions of radiotherapy. A routine follow-up on April 22 this year revealed no abnormalities except for a slightly low white blood cell count (WBC). The patient sleeps well (with the use of benzodiazepines), has one bowel movement per day, and normal urination. Past medical history includes low blood pressure, typically around 90/60 mmHg. |
| Presenting Symptoms | The patient currently experiences a sensation of heat in the nasal cavity and oral cavity, dryness in the nose and mouth, numbness of the scalp, a feeling of tightness or constriction in the head, stiffness in the neck and nape, coldness in the lower back and below, coldness in the epigastric region, dull pain in the epigastric area, fatigue, reduced appetite, one bowel movement per day with dry stool, a dark purple tongue, thin yellow coating, a greasy coating at the root of the tongue, and a deep, thin pulse. |
| Syndrome | Syndrome of Qi and Yin deficiency |
| Label Prescription | Radix Astragali (生黄芪) 20g, Radix Glehniae (北沙参) 20g, Rhizoma Atractylodis Macrocephalae (生白术) 15g, Poria (茯苓) 15g, Radix Angelicae Sinensis (当归) 10g, Fructus Ligustri Lucidi (女贞子) 15g, Herba Ecliptae (旱莲草) 10g, Spica Prunellae (夏枯草) 15g, Radix Paeoniae Alba (白芍) 30g, Radix Glycyrrhizae (甘草) 5g, Pericarpium Citri Reticulatae (陈皮) 12g, Bombyx Batryticatus (僵蚕) 10g, Herba Solani Nigri (龙葵) 20g, Herba Duchesneae Indicae (蛇莓) 15g, Radix Asparagi (天冬) 15g, Germinatus Oryzae (生稻芽) 15g |

| | Model predictions on the above TCM clinical record |
|---|---|
| Model Name | Tianyi |
| Q1: Analyze the current patient clinical record and provide the syndrome diagnosis and TCM prescription results. | **Syndrome Diagnosis**: Deficiency of Qi and Yin (气阴不足证). **Prescription**: Astragalus membranaceus (生黄芪) 20g, Pseudostellaria heterophylla (太子参) 20g, Radix Glehniae (北沙参) 20g, Ophiopogon japonicus (麦冬) 15g, Schisandra chinensis (五味子) 10g, Scrophularia ningpoensis (玄参) 20g, Trichosanthes kirilowii (天花粉) 15g, Coptis chinensis (黄连) 10g, Rehmannia glutinosa (生地) 20g, Angelica sinensis (当归) 15g, Paeonia lactiflora (白芍) 20g, Prunella vulgaris (夏枯草) 15g, Honey-fried Glycyrrhiza uralensis (炙甘草) 6g |
| Q2: Analyze the prescription you provided according with the perspective of the efficacy of traditional Chinese medicine and the composition theory of TCM prescription(Ju Chen Zuo Shi, also called by sovereign、minister、assistant and guide). | **Prescription analysis**: **Monarch Herbs**: Astragalus membranaceus (生黄芪), Pseudostellaria heterophylla (太子参), Radix Glehniae (北沙参), Ophiopogon japonicus (麦冬), Schisandra chinensis (五味子). Tonify Qi and nourish Yin. **Minister Herbs**: Scrophularia ningpoensis (玄参), Rehmannia glutinosa (生地), Angelica sinensis (当归), Paeonia lactiflora (白芍), Prunella vulgaris (夏枯草). Cool blood, resolve toxins, soften hardness, and dissipate nodules. **Assistant Herbs**: Coptis chinensis (黄连), Trichosanthes kirilowii (天花粉). Clear heat and resolve toxins. **Guide Herb**: Honey-fried Glycyrrhiza uralensis (炙甘草). Harmonize the actions of the other herbs. |
| Q3: Based on the syndrome diagnosis results you provided, analyze why the patient experienced these symptoms from the perspective of etiology. | **Etiology analysis in TCM**: The patient's current symptoms are partly caused by the side effects of treatments such as surgery, chemotherapy, and radiotherapy, and partly due to the progression of the disease itself. After surgery, the patient suffered damage to both Qi and blood, as well as impairment of the spleen and stomach, leading to dysfunction in transportation and transformation, insufficient production of Qi and blood, and symptoms such as fatigue, dull epigastric pain, poor appetite, dry stools, normal urination, and poor sleep. Postoperative chemotherapy and radiotherapy further damaged Qi and blood, resulting in fatigue and leukopenia. After radiotherapy, the patient developed radiation-induced pneumonia, radiation-induced esophagitis, and radiation-induced gastric ulcers. Disease progression led to lymph node metastasis, liver metastasis, and bone metastasis. Lymph node metastasis caused nerve compression, resulting in pain. Radiation-induced esophagitis caused difficulty in swallowing. Radiation-induced pneumonia caused dry cough with little phlegm. Chemotherapy and radiotherapy damaged Qi and blood, leading to Qi deficiency and failure to control blood, resulting in symptoms such as nosebleeds and gum bleeding. |

Fig. 3 Example of the predictions of Tianyi in TCM-SPHE. **a**. a TCM clinical Case. It consists of the medical history, current symptoms, the TCM syndrome diagnosis and the TCM prescription that provided by the TCM doctor. **b**. The prediction of Tianyi over the three continuous questions of this TCM clinical case in TCM-SPHE. The Tianyi take the medical history, current symptoms and the questions as the input, and generate the corresponding answers sequentially.

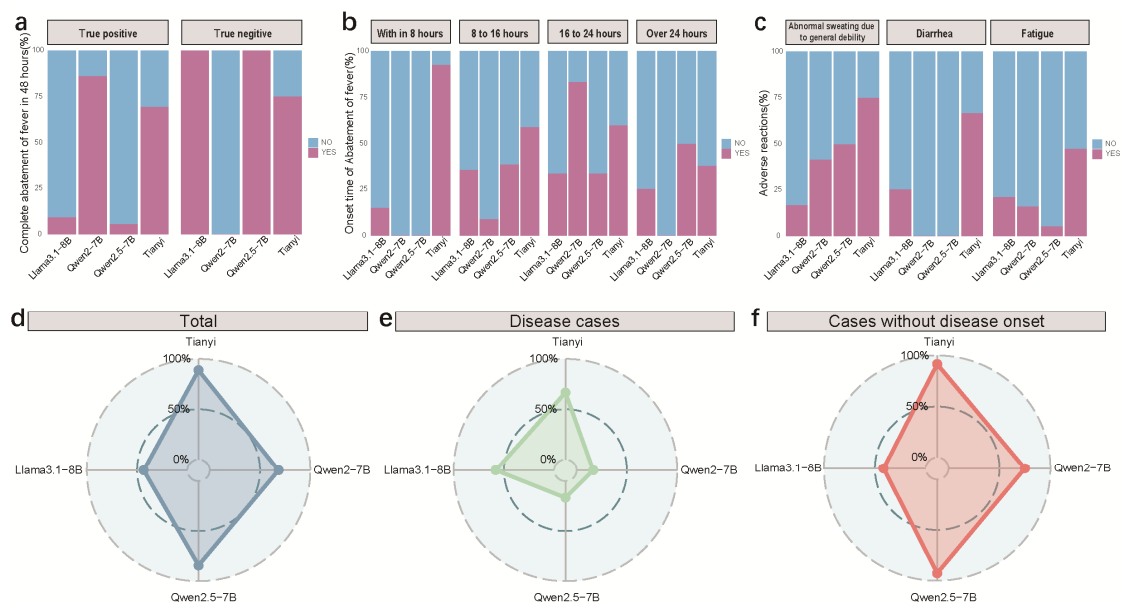

Fig. 4 **a**, **b**, **c** demonstrate the accuracy of the compared 4 LLMs for the 3 tasks (i.e., a. **complete abatement of fever in 48 hours, b. onset time of abatement of fever, c. adverse reactions**) of the clinical trial on the efficacy of traditional Chinese medicine for patients with H1N1 influenza for prospective study, where 'YES' means the portion that model correctly predicted and 'NO' means model's wrong predictions. The values outside the brackets are the LLM prediction while the values inside the brackets are the total number for that type. For **complete abatement of fever in 48 hours**, we mainly focus on 2 aspects: 1) the number of fever-reducing samples correctly predicted by the model, named as **True positive**; 2) the number of non-fever-reducing samples correctly predicted by the model, named as **True negative**. For **onset time of abatement of fever**, we divide the time prediction into 4 time-spans: [0, 8], (8, 16], (16, 24], >24. We evaluate the proportion of clinical records for which the LLM correctly predicts the time span during which the antipyretic effect occurs. For **adverse reactions**, we evaluate the proportion of clinical records for which the LLM correctly predicts if the target adverse reaction occurs. **d**, **e**, **f** are **the influenza incidence rate within 7 days after medication,** where **d** shows the total accuracy of model prediction, **e** presents the accuracy of correctly predicting disease onset, **f** demonstrates the accuracy of correctly predicting cases without disease onset. The accuracy of selected LLMs for retrospective study. We present the accuracy and the correct prediction numbers for each LLM. The value with percentage outside of the brackets and the number inside of the brackets are the accuracy and the number of correct predictions, respectively. The total number of disease cases and cases with disease are placed in the first column.

**Evaluations on Traditional Chinese Medicine Clinical Trial Prediction**

We select the LLMs with remarkable performance among the previous evaluations: Llama3.1-8B-instruct, Qwen2/2.5-7B-instruct, and Tianyi due to the consideration of the effectiveness and efficiency of prediction. As illustrated in panels **a**, **b,** and **c** of Fig. 4, the four LLMs demonstrate varying degrees of predictive capability across the three tasks. Their performance is significantly influenced by their comprehension of task requirements, clinical record information, and their respective medical knowledge bases.

Regarding the **complete abatement of fever within 48 hours**, Qwen2-7B-instruct and Qwen2.5-7B-instruct exhibit diametrically opposed results. Specifically, Qwen2-7B-instruct achieves near-perfect accuracy in predicting true positive cases but fails entirely in identifying true negatives. Conversely, Qwen2.5-7B-instruct demonstrates the exact opposite pattern. Analysis of the prediction outputs for all 112 valid clinical records reveals that Qwen2-7B-instruct predominantly predicts "Yes" with a few instances of "unpredictable," while Qwen2.5-7B-instruct uniformly predicts "No" across all records. Consequently, we conclude that these two models lack the capability to generate accurate predictions for this task based solely on task definitions and clinical record information.

In contrast, Llama3.1-8B-instruct demonstrates moderate predictive capability, correctly identifying 10 true positive cases and achieving perfect accuracy in true negative predictions. This suggests that the model can leverage specific clinical information and its acquired TCM knowledge to make predictions, albeit with limitations. Tianyi emerges as the most promising model for this task, correctly predicting approximately 75% of both true positive and true negative clinical records. This performance highlights the model's professional medical competence, particularly noteworthy given the extreme sparsity of true negative cases in the dataset. For the **onset time of fever abatement**, Qwen2/2.5-7B-instruct models show similar prediction patterns, achieving some accuracy within the [8, 16) and [16, 24) hour intervals. This performance mirrors their results in the 48-hour fever abatement task, indicating a strong bias toward specific outcome classes in both tasks. Based on these observations, we conclude that Qwen2/2.5-7B-instruct models are fundamentally incapable of making reliable clinical outcome predictions based on patient information and task requirements. Llama3.1-8B-instruct and Tianyi demonstrate more consistent and promising performance across all four time-intervals. Tianyi excels in predicting the correct onset time for most patients, while Llama3.1-8B-instruct achieves only limited success. Notably, all four models struggle with predictions for the "over 24 hours" interval, suggesting that without specific TCM clinical knowledge fine-tuning, LLMs tend to make overly optimistic predictions for fever reduction onset times. In the domain of **adverse reaction** prediction, Tianyi demonstrates superior accuracy for the first two adverse reaction types among the four models, though its performance on fatigue prediction is relatively weaker. Qwen2/2.5-7B-instruct models show similar performance patterns to their results in the previous two tasks. Llama3.1-8B-instruct maintains its limited but superior predictive capability compared to Qwen2/2.5-instruct models in this task.

Regarding **the influenza incidence rate within 7 days post-medication**, the performance of these four LLMs aligns with the clinical trials depicted in panels **a**, **b**, and **c** of Fig. 4. Qwen2/2.5-instruct models demonstrate competitive accuracy in predicting cases without disease onset but exhibit minimal predictive success for actual disease cases. This pattern mirrors their performance in **the complete abatement of fever within 48 hours**, as shown in Fig. 4, further substantiating their limitations in clinical outcome prediction. In contrast, Tianyi and Llama3.1 outperform their results for this trial. Upon analyzing the discrepancies between these two trials, we observe that the medication data in this trial reflect actual patient adherence, which varies significantly - some patients fully comply with the prescribed TCM regimen, while others do not. Conversely, the H1N1 trial assumes uniform patient adherence: "patients take the entire dose of the TCM prescription as prescribed," which fails to capture real-world medication variability.

Based on the evaluation of these two clinical trials, we can conclude that Tianyi and Llama3.1-8B-instruct demonstrate remarkable performance and potential in TCM-related decision-making capabilities (including syndrome-prescription prediction, explanation, and etiology-pathogenesis analysis) and clinical outcome prediction, even without explicit instruction fine-tuning. Llama3.1-8B-instruct achieves performance comparable to Tianyi in TCM syndrome-prescription tasks but falls short of desired accuracy in TCM clinical trials. For syndrome-prescription tasks, LLMs can deliver satisfactory results by mastering TCM diagnostic standards and herbal medicine applications, logically applying this knowledge to patient symptoms and physical signs. In other words, success in these tasks primarily requires memorization and proper application of TCM syndrome and medicine knowledge. However, for clinical trials, LLMs must not only possess comprehensive and in-depth understanding of TCM knowledge but also demonstrate mastery of clinical practice experience. To effectively integrate LLMs into professional medical fields or current clinical workflows, they must be deliberately designed and fine-tuned to meet specific TCM mission requirements. Tianyi, despite not being thoroughly and comprehensively fine-tuned for all TCM missions, demonstrates significant potential in medical applications.

**Discussion**

In this work, we present Tianyi, a scale-proper large language model for the learning and application of natural medicines and the theoretical diagnosis-treatment knowledge based on it (i.e., Traditional Chinese Medicine knowledge system), which is trained and fine-tuned in a comprehensive manner to simulate the TCM clinician expertise across the TCM general disciplines and various application scenarios. Evaluated on the TCMEval, Tianyi emerges the superiority among a wide-range of TCM specialties and tasks. Besides, the extensive comparison studies with the same model scale indicates its potential of real-world clinical case assistance diagnosed and treated by TCM for the junior doctors with limited clinical experience in diagnosis and treatment. Furthermore, the real-world evaluations on TCM real clinical cases and clinical trial studies demonstrate the capability on solving the unseen, real-world clinical tasks, indicating Tianyi's potential as a top-tier TCM assistant for clinician in their daily works.

  The TCM knowledge system, as a holistic medical framework, is designed to differentiate and diagnose patient symptoms and etiologies, and to treat them using a variety of natural medicines. This system is inherently complex, challenging to master thoroughly, and difficult to teach and pass down through generations. It transcends the simplistic correlation between a patient's symptoms and the corresponding natural remedies. Instead, it necessitates a comprehensive analysis of the underlying causes, accurate diagnosis of TCM syndromes, formulation of treatment principles, and the prescription of TCM formulas composed of natural medicines. This intricate process requires the integration of knowledge from multiple TCM disciplines, including TCM Diagnostics, TCM Internal Medicine or Surgery, Pharmacology of Chinese Medicines, TCM Prescription Studies, and TCM Pharmacology, among others. The sophisticated application of such diverse medical knowledge is prone to errors, often resulting in misdiagnosis and inappropriate treatment. Previous AI-assisted systems for TCM have included rule-based clinical decision-making systems, expert systems, and machine learning or deep learning models trained on Electronic Health Records (EHRs). However, these systems are constrained by their limited model scale and the biased, oversimplified nature of their knowledge bases.

The recent researches of LLM demonstrate its advanced potential in understanding, interpreting, and generating the text effectively across diverse conditions and requirements with the minimal specific fine-tuning[35]. However, these advancements fall short when applied to complex and diverse clinical scenarios, particularly in the context of TCM. Existing LLMs are not capable of efficiently modeling the vast of TCM medical knowledge in a scalable manner. Furthermore, the special designed TCM LLMs of similar scale to Tianyi are mainly focus on the chatbot of TCM consultation purpose, trained and fine-tuned by using limited TCM resources and instruction samples. This results in the lack of systematic TCM knowledge, factual inaccuracies, logic inconsistencies, and an inability to handle TCM clinical tasks. For instance, the widely cited Zhongjing-Llama, despite being trained and fine-tuned on a broad range of TCM corpora and instruction samples for clinical consultation purposes, fail to properly perform the diverse clinical tasks we designed in this work no matter of the basic ones and the real-world evaluations[14]. To address these challenges, we introduce Tianyi, which, to our knowledge, is the first TCM LLM that has been comprehensively trained and fine-tuned on diverse TCM corpora. It is designed with advancements in model scalability, outstanding clinical capabilities, and adaptability to various clinical scenarios. In our viewpoint, for the development of medical LLM, we should not solely focus on increasing the model scale and/or exploiting medical resources in a superficial manner. Instead, the specialized and professionally designed strategy should be adopted, tailored to the diverse medical conditions and considering cost-effectiveness for real-world medical practice, where the sufficient computational resources are scarce to deploy the LLMs with very large model scale. While Tianyi demonstrates the promising and advanced TCM clinical performance with a model scale approximately 10% that of the comparable LLM, alleviate these challenges and offering the insights for developing medical LLM assistant for clinicians.

We conduct a comprehensive basic and clinical evaluations for the popular LLMs and Tianyi within practical clinical scenarios, which includes the **Regular TCM Benchmark Tasks** and the **TCM Real-World Evaluation**. In **Regular TCM Benchmark Tasks**, we built three TCM benchmarks: TCM multiple choice exams, TCM domain question answering, and TCM clinical records tasks. The first two benchmarks are the factual knowledge evaluations that include general TCM disciplines which cover a wide array of aspects, assessing the LLM's proficiency in TCM knowledge. The third benchmark contains a collection of historical TCM case records, of which each record consists of three tasks: the TCM syndrome diagnosis, the treatment principle decision, and the TCM prescription generation. The LLM need to provide three predictions at first. Then, the average F1-score is calculated based on three tasks' F1-scores. This task with increasing difficulty valid the extent of solving the TCM essential tasks by applying the TCM knowledge in the comprehensive manner. As shown in Table 1, results demonstrate that our model is not only learnt professional TCM knowledge capacity to complete the ordinary knowledge problems, but also capable of solving the syndrome diagnosis, treatment principle decision, and TCM prescription at the same time, demonstrating its potential of being the clinician's assistance in the TCM clinical practice. Additionally, we conduct the TCM clinical study on the prediction of clinical trial researches, which are the scenarios that all LLMs including Tianyi are unseen before. It evaluates the extent of LLMs' ability emergence, and valid if the LLM can induce and deduce the knowledge they learned, analyze the clinical condition, obey the task instructions to provide the reliable predictions for these clinical trial researches. All LLMs predict the results for each case of the clinical trial by converting the original clinical indexes of symptoms, physical signs, medications into the designed prompt by zero-shot learning. The results show the promising potential of Tianyi to assist the researchers in the natural medicines efficacy for different patients and adverse reactions prediction. Such ability demonstrates the probable of enable LLM prediction ability in various clinical study tasks if the knowledge and experience of target clinical study can be designed for LLM learning, enhancing the efficiency and effectiveness of clinical study.

Although our model has demonstrated the promising performance, several challenges remain. First, as for the proper use of natural medicines, TCM doctors use observation, listening and smelling, inquiry, and palpation (called, the four diagnostic methods of TCM) to gather comprehensive information about a patient's condition. The observation and palpation are the different modalities from text. How to integrated the multi-modal data into the process of model's information reasoning is still an open theme. In the future, studying and integrating multi-modal data and approaches could enable Tianyi to adopt a more comprehensive way to improve its performance on diverse TCM clinical scenarios from the raw multi-modal settings, enhancing the human–computer collaboration. Furthermore, improving Tianyi's complex reasoning capabilities could significantly enhance its performance and applicability in both clinical practice and in-depth research. Currently, while Tianyi possesses reasoning abilities, these were not explicitly designed and remain at a foundational level. There is considerable potential for Tianyi to advance its deductive capabilities in areas such as the clinical analysis of TCM syndromes and etiologies, the rationale behind treatment principles, the selection of natural medicines, and the logical progression of illnesses and corresponding adjustments in natural medicine prescriptions. Strengthening Tianyi's reasoning abilities would not only improve its practical integration into clinical workflows but also maximize its potential to benefit healthcare practices, particularly in primary care settings.

**Data availability**

Due to privacy regulations, the pretraining corpora and EHRs cannot be made freely available in a public repository. De-identified data from TCMEval can be requested through the management team by contacting the corresponding author (C.Y.L), following a defined protocol for data request approval. Generally, all such requests for access to data will be responded to within 1 month. We will release our model publicly available for the research purpose.

## Competing interests
We declare that none of the authors have competing financial or non-financial interests as defined by Nature Portfolio.

## Acknowledgements
This study was funded by the National Key Research and Development Program (grant 2023YFC2308200), the National Natural Science Foundation of China (grants 82205317, and 82405609), the Fundamental Research Funds for the Central Universities (2024-JYB-JBZD-064), National High Level Chinese Medicine Hospital Clinical Research Funding(K2023C14). The authors would like to thank physicians (Y. Mao, S. Liu, L. Li) from the Dongfang Hospital Affiliated to Beijing University of Chinese Medicine, physicians (C. Bai and D. Dou) from the Beijing University of Chinese Medicine, physician (Y. Wu) from Shenzhen Hospital Affiliated to Beijing University of Chinese Medicine, physician (J. Wu) from Hubei Provincial Traditional Chinese Medicine Hospital, physician (D. Wang) from Beijing Gulou Traditional Chinese Medicine Hospital, physician (Y. Wang) from Shunyi Campus, Beijing Hospital of Traditional Chinese Medicine, for assistance on the real-world evaluations.

## Author contributions
Z.L., T.Y., C.Y.L., J.X.Y. and B.J.L. conceived and designed the methodology and analyzed the data. Z.L., B.J.L., X.C.L., K. C., J.W., and Z.G. collected and processed the data. P.C., X.H.G., Y.L., and Y.X.C. supervised the project. Z.L., B.J.L., J.X.Y., K. C., and X.C.L. trained, fine-tuned, and evaluated the models. All authors discussed the results and approved the paper.

**Correspondence and requests for materials** should be addressed to Peng Cao, Changyong Luo, Xiaohong Gu, or Yan Li.